\documentclass{article}
\usepackage{spconf,amsmath,graphicx}

\usepackage{times}  % DO NOT CHANGE THIS
\usepackage{helvet}  % DO NOT CHANGE THIS
\usepackage{courier}  % DO NOT CHANGE THIS
\usepackage[hyphens]{url}  % DO NOT CHANGE THIS
\usepackage{graphicx} % DO NOT CHANGE THIS
\usepackage{caption} % DO NOT CHANGE THIS AND DO NOT ADD ANY OPTIONS TO IT
\usepackage{algorithm}
\usepackage{algorithmic}

\usepackage{makecell}
\usepackage{booktabs}
\usepackage{multirow}
\usepackage{amsmath}
\usepackage{graphicx}
\usepackage{float}

\usepackage[capitalize]{cleveref}
\crefname{section}{Sec.}{Secs.}
\Crefname{section}{Section}{Sections}
\Crefname{table}{Table}{Tables}
\crefname{table}{Tab.}{Tabs.}

\usepackage{newfloat}
\usepackage{listings}

\title{Filter Pruning via Filters Similarity in Consecutive Layers}

\name{Xiaorui Wang$^{1}$, Jun Wang$^{1\ast}$, Xin Tang$^{2}$, Peng Gao$^{1}$, Rui Fang$^{2}$, Guotong Xie$^{1}$\sthanks{Corresponding authors. \{wangjun916, xieguotong\}@pingan.com.cn.}}

\address{$^{1}$AI Platform Group, Ping An Technology Co. Ltd., Beijing, China \\
$^{2}$Visual Computing Group, Ping An Property $\&$ Casualty Insurance Company, Shenzhen, China}

\begin{document}
%\ninept
%
\maketitle
\begin{abstract}
Filter pruning is widely adopted to compress and accelerate the Convolutional Neural Networks (CNNs), but most previous works ignore the relationship between filters and channels in different layers. Processing each layer independently fails to utilize the collaborative relationship across layers. In this paper, we intuitively propose a novel pruning method by explicitly leveraging the Filters Similarity in Consecutive Layers (FSCL). FSCL compresses models by pruning filters whose corresponding features are more worthless in the model. The extensive experiments demonstrate the effectiveness of FSCL, and it yields remarkable improvement over state-of-the-art on accuracy, FLOPs and parameter reduction on several benchmark models and datasets. 
\end{abstract}

\section{Introduction}
\label{sec:intro}

Recently, the large model size and high computational costs remain great obstacles for the deployment of CNN models on devices with limited resources. Model compression, which can reduce the sizes of networks, 
mainly falls into some categories, i.e., pruning, quantization, knowledge distillation~\cite{he2018soft}.
Pruning methods can handle any kind of CNNs and will not have substantial negative influences on model performance. Specifically, Typical pruning contains: weight pruning and filter (channel) pruning~\cite{he2018soft, he2019filter}. Weight pruning directly deletes weight values in a filter in an irregular and random way, and may cause unstructured sparsities. So it is unable to achieve acceleration on general-purpose processors~\cite{he2018soft}. Meanwhile, filter (channel) pruning discards the whole selected filters, thus the pruned network structure 
won't be damaged and can easily achieve acceleration in general processors~\cite{he2019filter}.

\begin{figure}[!t]
	\centering
	\includegraphics[width=0.9\columnwidth]{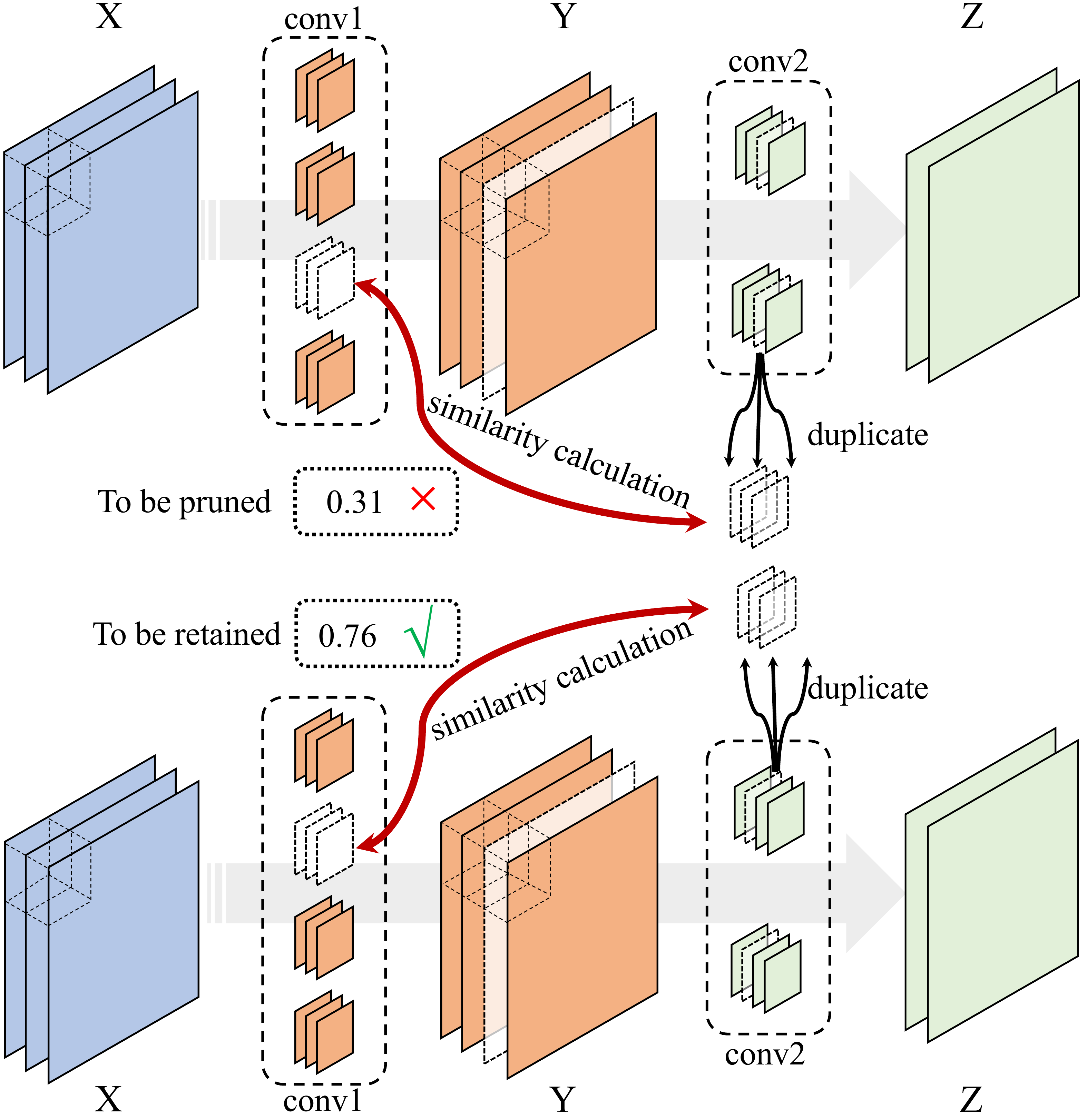}
	\caption{Filter pruning via Filters Similarity in Consecutive Layers (FSCL). There are 4 and 2 filters at the 1st and 2nd layers, respectively. If the 3rd filter in conv1 is pruned, the 3rd feature map can be removed, implying that the 3rd channels of the 2 filters at conv2 can be removed, too. On the top, FSCL calculates the similarity between the 3rd filter at conv1 and the 3rd channels of the 2 filters at conv2 to evaluate the importance of the 3rd filter at conv1. On the bottom, we similarly evaluate the importance of the 2nd filter at conv1. Then we prune the 3rd filter for its smaller importance score.}
	\label{fig:FSCL}
	\vspace{-0.5cm}
\end{figure}

Current filter pruning can be implemented in filter-wise or channel-wise manner. Most of them consider the information in each independent convolution layer, and they do not use the relationship between filter and channel in more layers explicitly. When evaluating the importance of filters, different layers usually cannot communicate with each other. And most of them conceive that the norm of a filter and its importance are strongly correlated. But this has not been proved by theory. To utilize this criterion, analysis in FPGM~\cite{he2019filter} shows that some prerequisites should be satisfied. And experiments in Taylor~\cite{molchanov2019taylor} show that there is a significant gap in the correlation between the norm and the importance.

To address the above limitations, we propose a novel filter pruning via Filters Similarity in Consecutive Layers (FSCL), to merge the information in two continuous layers for evaluation of filter importance. FSCL calculates the similarity of filters in two consecutive layers by convolution operation, and it can quantify the worth of features extracted by the filters. Then we prune the filters 
tend to have little contribution to the network. FSCL not only takes the filters which generate the feature maps into consideration, but also takes advantage of the channels in the next layer which uses the feature maps.

We highlight the main contributions as follows:

\noindent(1) We explore the relationship between filter-wise and channel-wise pruning, reveal reasonable information in continuous convolutional layers that can be used to prune filters.

\noindent(2) We propose a novel method to estimate the contribution of a filter in each convolution layer using the Filter's Similarity in Consecutive Layers (FSCL), which combines filter-wise pruning with channel-wise pruning.

\noindent(3) Experimental results show that the proposed FSCL achieves state-of-the-art performance on a wide variety of networks trained on CIFAR-10 and ImageNet.

% methods
\section{Methodology}

%% problem definition setup notations
This section explains FSCL by performing filter pruning on CNNs. 
We evaluate the importance (similarity) of filters in convolutions by two consecutive layers, as shown in Figure~\ref{fig:FSCL}.

\subsection{Preliminary}
% \subsection{Filter Pruning via Importance}

We assume that a neural network has $L$ convolutional layers, and $W^{i}$ is the parameters in the $i$-th convolutional layer. We use $N^{i}$ to represent the number of filters in $W^{i}$. The parameters $W^{i}$ can be represented as a set of 3D filters $W^{i}=\left \{ w^{i}_{1}, w^{i}_{2}, ... , w^{i}_{N^{i}} \right \} \in R^{N^{i} \times C^{i} \times K^{i} \times K^{i}}$. $C^{i}$ is the number of channels in filters and $K^{i}$ denotes the kernel size.

%evaluate filter importance and
Filter pruning is to prune the filters from $N^{i}$ to desired $N^{i}_{0} (0 \le N^{i}_{0} \le N^{i})$. The core is to remove the less important filters, which can be formulated as an optimization problem:

\begin{equation}\label{filter_importance}
\begin{aligned}
&\mathop{\arg\max}\limits_{\beta_{i,j}}\sum_{i=1}^{L}\sum_{j=1}^{N^{i}}\beta_{i,j}\mathcal{L}({w}_j^i), \\
&\begin{array}
{r@{\quad}@{}l@{\quad}l}
s.t. &\sum_{j=1}^{N^{i}}\beta_{i,j}& \leq N^{i}_{0}, \\
\end{array}
\end{aligned}
\end{equation}
where $\beta_{i,j}$ is an indicator which is 1 if ${w}_j^i$ is to be reserved, or 0 if ${w}_j^i$ is to be removed. $\mathcal{L}(\cdot)$ measures the importance of a filter. 
 
Designing $\mathcal{L}(\cdot)$ has been widely studied in the community~\cite{li2017pruning,he2019filter}. 
L1~\cite{li2017pruning} measures the importance of each filter by calculating its absolute weight sum. FPGM~\cite{he2019filter} calculates the distance to the geometric median of filters as filter importance score. Unfortunately, these above definitions ignored the relationship between consecutive layers.

\subsection{Filters Similarity in Consecutive Layers}
\label{sec:FiltersSCL}

We propose to define $\mathcal{L}$ on the worth of features extracted by the filters. The worth is proportional to the usefulness of the features. 
To calculate the usefulness of features, we firstly look into the convolution operation in consecutive layers and the filter pruning in them. % As shown 
In Figure~\ref{fig:FSCL}, we randomly sample an element in the feature map $Y$ of the $c$-th ($c$ is short for current) layer, it is in the $j^{c}$ channel and denote as $y_{j^{c}}$, $j^{c} \in \{ 1, 2, \dots , N^{c}\}$. A corresponding filter $w_{j^{c} } \in R^{C^{c} \times K^{c} \times K^{c}}$ and sliding window  $x \in R^{C^{c} \times K^{c} \times K^{c}}$ can also be determined according to its location. The convolution operation is: %computed as follows:
\begin{equation}
    y_{j^{c} } =\sum_{c^{c}=1}^{C^{^{c} } } \sum_{k^{^{c} }_{1}=1}^{K^{c} } \sum_{k^{^{c}}_{2}=1}^{K^{c} }w_{j^{c},c^{c},k_{1}^{c},k_{2}^{c}}^{c}\times x_{c^{c},k_{1}^{c},k_{2}^{c}}.
\end{equation}
%For simple representation, bias term is not included in our formulation. 
Similarly, a  randomly sampled element $z_{j^{n}}$ in the $j^{n}$th ($n$ is short for next and $j^{n} \in \{ 1, 2, \dots , N^{n}\}$) channel of next layer's feature map $Z$  is computed as follows:
\begin{equation}
    z_{j^{ n} } =\sum_{c^{ n}=1}^{C^{^{ n} } } \sum_{k^{^{ n} }_{1}=1}^{K^{ n} } \sum_{k^{^{ n}}_{2}=1}^{K^{ n} }w_{j^{ n},c^{ n},k_{1}^{ n},k_{2}^{ n}}^{ n}\times y_{c^{ n},k_{1}^{ n},k_{2}^{ n}}.
\end{equation}

%As shown 
In Figure~\ref{fig:FSCL}, if we remove the $j_{0}^{c}$th filter $w_{j_{0}^{c},:,:,:}^{c}$, the $j_{0}^{c}$th  channel of feature map $Y$ is close to zero, implying that the  $j_{0}^{c}$th  input channel of the $N^{n}$ filters in next convolution layer are prone to be useless. The  $j_{0}^{c}$th channel of the convolution $w_{:,j_{0}^{c},:,:}^{n}$ can be removed too. The filter number of the current layer $N^{c}$ are the same as the channel number of the next layer $C^{n}$.
So we find when evaluate the usefulness of the features extracted by the filters, we should consider two parts: The filter that products it and the channel which uses it. The second part is very important but ignored by other methods. Only the features used for the subsequent calculations are valuable.

To define $\mathcal{L}$ on the worth of features extracted by the filters, we evaluate the similarity in continuous layers. %The rationale lies in that t
The more similar they are, the more features extracted by previous filters will be used for the next layer, and finally, for the prediction.%ed labels.

The convolution operation in layer $n$ is calculated by:
\begin{equation}
    w_{:,j_{0}^{c},:,:}^{n} \times y_{j_{0}^{c},:,:},
\end{equation}
but %when we don't have the value in 
without feature map $Y$, we can use the parameters in the filter which product it to similarly replace it in the calculation. For dimensional consistency, we duplicate the $j^{n}$th filter's $j_{0}^{c}$th channel $w_{j^{n},j_{0}^{c},:,:}^{n}$  $C^{c}$ times and concatenate them together, then we get a new convolution filters $\hat{w}_{j^{n},j_{0}^{c},:,:,:}^{n}$. $j^{n} \in \{ 1, 2, \dots , N^{n} \}$, so we can similarly get $N^{n}$ new convolution filters. 
Filters extract features most similar to it. We can calculate the similarity between the new convolution filters and the filter in the previous layer $w_{j_{0}^{c},:,:,:}^{c}$ by convolution operation. 
The result can be used as an importance score for model pruning. We average the sum of the absolute value of $N^{n}$ results as the importance score of the $j_{0}^{c}$th filter $w_{j_{0}^{c},:,:,:}^{c}$:
\begin{equation}
    \mathcal{L}({w}_{j_{0}^{c}}^c)=\frac{1}{N^n } \sum_{j^{n}=1}^{N^n} \left \| {w}_{j_{0}^{c},:,:,:}^c  \otimes  \hat{w}_{j^{n},j_{0}^{c},:,:,:}^{n} \right \|_{1},
\end{equation}
where $\otimes$ denotes the convolution operation and $\left \|  \cdot \right \|_{1}$ denotes $\ell_1$-norm. 
Instead of  considering a single layer, our proposed method evaluates the channels with respect to other filters consecutively placed in the next layer. 
A similarity measure calculated between filters would determine if the filters should be retained, where a high similarity points out more information being extracted, so the corresponding filters are more important.
Our FSCL can measure the worth of features extracted by the filters, % effectively, 
without being affected by the distribution of the input. Besides, we can calculate the importance scores offline. % What's more, c
Compared with FPGM~\cite{he2019filter}, which calculates difference between filters in one convolution, our method calculates difference between filters in continuous convolutions.
After the definition of $\mathcal{L}$, Eq.\,(\ref{filter_importance}) can be solved by pruning the filters with $(N^{i}-N^{i}_{0})$ least importance scores.  
Then we fine-tune the pruned model to restore accuracy.

\vspace{-0.2cm}

\subsection{FSCL For Multiple Structures}

We provide standards for FSCL in different structures include: Plain structure, Inception module, Residual block.

\noindent\textbf{Plain Structure}.
For the Plain structure, such as VGG, we use vanilla FSCL to evaluate the importance of filters.

%\paragraph{GoogLeNet}
\noindent\textbf{Inception Module}.
There are four branches in inception modules of GoogleNet: branch1x1, branch3x3, branch5x5, branchpool. The feature maps and the next layers of them are concatenated together. To implement FSCL in GoogleNet, we first split the channels of the next layers by different branches, thus the four first layers of the next inception module are all split into four groups. Take branch1x1 as a example, it links to four branches, so we concatenate the channels in four first layers of the next inception module together as a new filter.

%\paragraph{ResNet-50}
\noindent\textbf{Residual Block}.
(1) \textbf{For the residual block with no convolution in shortcut}, such as ResNet-56, we use the method explained in Plain Structure. (2) \textbf{For residual blocks with convolution in shortcut}, such as Resnet-50, we multiply the importance scores of shortcut connections and residual connections together to get global importance scores and then prune them together.

\vspace{-0.1cm}

\section{Experiments}
\label{sec:experiment}

\subsection{Implementation Details}

\textbf{Datasets and Models}.
We evaluated our method on CIFAR-10 and ImageNet. Various recent network architectures were used in our experiments.%, i.e., 

\noindent\textbf{Configurations}.
All experiments for CIFAR-10 and ImageNet were conducted on 1 and 8 NVIDIA V100 GPUs, respectively. We first evaluated the filter importance by FSCL. Then we pruned the filters with less importance in each layer. Finally, we fine-tuned the slimmed network.
For comparison, we first fixed similar parameters and FLOPs, and measured the accuracy, then fixed the accuracy to be similar to the baselines and measured the parameters and FLOPs reductions.

\subsection{Comparisons of Accuracy}

\begin{figure}
	\begin{center}
		\centerline{\includegraphics[width=0.8\linewidth]{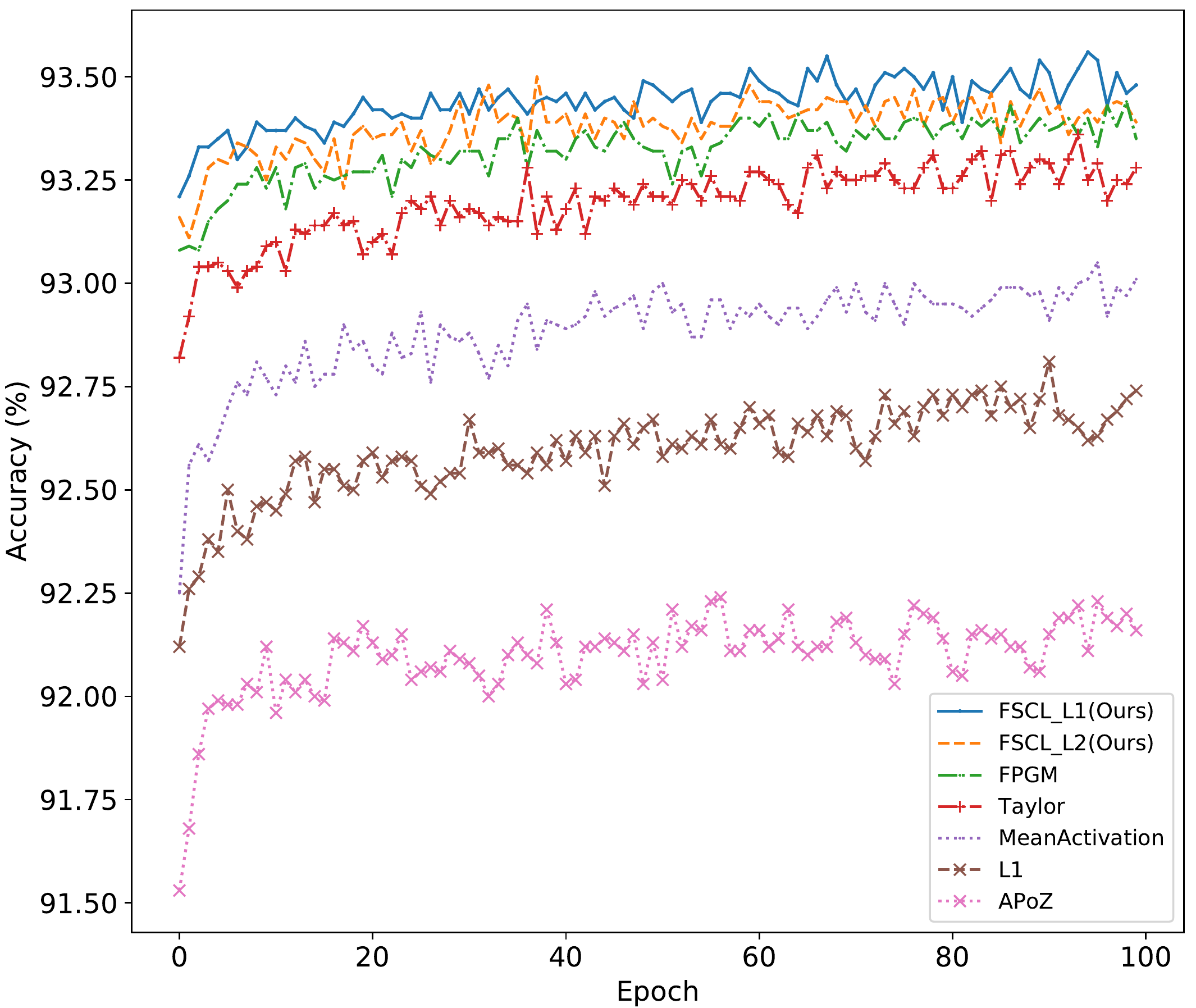}}
		\caption{The accuracy and epochs of pruned vgg-16 on cifar10 by different methods. Our FSCL\_L1 and FSCL\_L2 are consistently better and FSCL\_L1 yields best performance.}
		\label{fig:compare_log}
	\end{center}
\vskip -0.25in
\end{figure}

First, to evaluate of the accuracy under fixed similar reductions, we compared 
FSCL with recently proposed methods (Figure~\ref{fig:compare_log}). The L1 version of FSCL used the $\ell_1$-norm and the L2 version used the $\ell_2$-norm. Other baselines include: FPGM~\cite{he2019filter}, Taylor~\cite{molchanov2019taylor}, MeanActivation~\cite{DBLP:journals/corr/MolchanovTKAK16}, L1~\cite{li2017pruning} and APoZ~\cite{Hu2016NetworkTA}. None of these five methods considered the relationship between layers. We pruned the VGG-16 which accuracy was 93.99 by different filter pruning methods and fine-tuned the pruned model for 100 epochs on CIFAR-10. We used the implementations of these methods by NNI (https://github.com/microsoft/nni).

We pruned some of the convolution layers in VGG-16 (feature.0, feature.24, feature.27, feature.30, feature.34 and feature.37) by 50\% using different methods. 
Figure~\ref{fig:compare_log} shows the accuracy and epochs of fine-tune process. The L1 and L2 version of FSCL (“FSCL\_L1”, “FSCL\_L2”) achieved remarkable higher accuracy compared with other methods without considering the relationship between filters and channels in consecutive layers. It indicated that merging the weight information of consecutive layers was effective for the importance calculation of the convolution kernel. 
The L1 version was higher than L2, showing the L1 norm was a better criterion.

\subsection{Comparisons of Parameters and FLOPs}

Second, we evaluated whether FSCL can outperform baselines in reducing parameters under similar accuracy.
For VGGNet on CIFAR-10, we fine-tuned the pruned network for 300 epochs, with a start learning rate of 0.009, weight decay 0.005, and momentum 0.9. The learning rate is divided by 10 at the epochs 150, 225, 265. For ResNet-56 on CIFAR-10, we fine-tuned with a start learning rate of 0.01. The other settings are the same as VGGNet. 
For GoogLeNet on CIFAR-10, the training schedule was the same as VGGNet except for the start learning rate 0.008. After pruning ResNet-50 on ImageNet, we fine-tuned the pruned model for 220 epochs with a start learning rate of 0.009, weight decay 0.0001, and momentum 0.99. The learning rate schedule type is cosine annealing.

\begin{table}[!t]
	\caption{Comparisons of performance on CIFAR-10. "-" denotes that the results are not reported. Top-1 $\downarrow$ is the Top-1 accuracy gap between the baseline model and the pruned model. The best performance is highlighted in bold.}
	\label{tab:comparision_cifar10}
	\centering
	\scalebox{0.7}{
		\begin{tabular}{c|c|c|rr}
			\hline
			 & Methods &B/P Top-1(\%)/$\downarrow$ (\%) &FLOPs/$\downarrow$(\%) &Params/$\downarrow$(\%) \\
			\hline
			%\multirow{11}{*}{\tabincell{c}{ResNet-56}}
			\multirow{5}{*}{\rotatebox{90}{VGG-16}}
			&SSS~\cite{SSS2018}  &93.59  /  93.02 / 0.57 &183.13M / 41.6 & 3.93M / 73.8 \\\
			&GAL-0.1~\cite{Lin2019TowardsOS}  &93.96 / 93.42 / 0.54 &171.89M / 45.2 & 2.67M / 82.2 \\\
			&Hinge~\cite{Li2020GroupST}  &{94.02} / 93.59 / 0.43 & -\,\,\,\,\,\,\,\, / 39.1 & -\,\,\,\,\,\, / 80.1 \\\
% 			&HRankPlus~\cite{lin2020hrank} &93.96 / 93.10 &0.86 &66.95M / 78.6 &1.90M / 87.3   \\\
			&HRank~\cite{lin2020hrank} &93.96 / 92.34 / 1.62 &108.61M / 65.3 &2.64M / 82.1   \\\
			&\textbf{FSCL(Ours)} &93.96 / \textbf{93.68} / \textbf{0.28} &\textbf{58.06M / 81.5} &\textbf{1.58M / 89.5}  \\
			\hline
			\multirow{10}{*}{\rotatebox{90}{ResNet-56}}
% 			&L1~\cite{li2017pruning}  &93.04 / 93.06 &-0.02 &90.90M / 27.6 & 0.73M / 14.1 \\\
			&CP~\cite{He2017ChannelPF}  & 92.80 / 91.80 / 1.00 &62.00M / 50.6 & -\,\,\,\,\,\,\, / \,\,\, - \,\,\,\\\
			&NISP~\cite{Yu2018NISPPN}  & 93.04 / 93.01 / 0.03 &81.00M / 35.5 & 0.49M / 42.4 \\\
			&GAL-0.6~\cite{Lin2019TowardsOS}  & 93.26 / 93.38 / -0.12 &78.30M / 37.6 & 0.75M / 11.8 \\\
% 			&HRankPlus~\cite{lin2020hrank} &93.26 / 93.57 &-0.31 &65.94M / 47.4  &0.48M / 42.8   \\\
			&HRank~\cite{lin2020hrank} &93.26 / 93.17 / 0.09 &62.72M / 50.0  &0.49M / 42.4   \\\
			&NPPM~\cite{Gao_2021_CVPR} &93.04 / 93.40 / -0.36	&-\,\,\,\,\,\,\,\, / 50.0 &\,\,\,\,\,\,\,\,\,\,\,\,\,\,-\,\,\,\,\,\, / \,\,\, - \,\,\,   \\\
			&LRPET~\cite{guo2022compact} &93.33 / 93.10 / 0.23 &61.92M / 51.0  &0.43M / 49.4   \\\
			&\textbf{FSCL(Ours)} &{93.26 / \textbf{93.65}} / \textbf{-0.39	}	&\textbf{59.95M / 52.2} &\textbf{0.42M / 50.3}   \\
			\cline{2-5}
			&FPGM~\cite{he2019filter} &{93.59} / 93.49 / 0.10 	&-\,\,\,\,\,\,\,\, / 52.6  & -\,\,\,\,\,\,\, / \,\,\, - \,\,\, \\\
			&P Criterion~\cite{9156434} &{93.59} / 93.24 / 0.35 	&-\,\,\,\,\,\,\,\, / 52.6  & -\,\,\,\,\,\,\, / \,\,\, - \,\,\,  \\\
			&\textbf{FSCL(Ours)} &93.26 / \textbf{93.52}   / \textbf{-0.26}		&\textbf{57.26M / 54.4} &\textbf{0.43M / 49.7}   \\
            \hline
			\multirow{8}{*}{\rotatebox{90}{GoogLeNet}}
			&Random &95.05 / 94.54   / 0.51 &960M / 36.8 & 3.58M / 41.8 \\\
			&GAL-ApoZ~\cite{Hu2016NetworkTA} &95.05 / 92.11 / 2.94 &760M / 50.0 &2.85M / 53.7 \\\
% 			&GAL-L1~\cite{li2017pruning} &95.05 / 94.54  &0.51 &1020M / 32.9 & 3.51M / 42.9 \\\
			&GAL-0.5~\cite{Lin2019TowardsOS} &95.05 / 94.56  / 0.49 &940M / 38.2 &3.12M / 49.3 \\\
			
			&ABCPruner~\cite{ijcai2020-94} &95.05 / 94.84 / 0.21	&513M / 66.6 &2.46M / 60.1   \\\
% 			&HRankPlus~\cite{lin2020hrank} &95.05 / 94.82 &0.23	&395M / 73.9 &2.09M / 65.8   \\\
			&HRank~\cite{lin2020hrank} &95.05 / 94.53 / 0.52	&690M / 54.9 &2.74M / 55.4   \\\
			&DCFF~\cite{lin2021training} &95.05 / 94.92 / 0.13	&460M / 70.1 &2.08M / 66.3   \\\
			&CLR-0.91~\cite{Lin2022PruningNW}  &95.05 / 94.85 / 0.20	&490M / 67.9 &2.18M / 64.7   \\\
			&\textbf{FSCL(Ours)} &{95.05 / \textbf{95.03}}  / \textbf{0.02} &\textbf{375M / 75.3} &\textbf{2.07M / 66.3}   \\
			%\textbf{66.12} \\
			\hline
	    \end{tabular}
	} 

\end{table}

\noindent\textbf{VGGNet on CIFAR-10}.
As shown in Table~\ref{tab:comparision_cifar10}, 
FSCL achieves state-of-the-art performance. FSCL demonstrated its ability to obtain the lowest accuracy drop of 0.28\%, with 81.5\% FLOPs reduction and 89.5\% parameters reduction. And the pruned model has the highest Top-1 accuracy (93.68\%). 
FSCL utilizes the relationship between layers, which is the main cause of its superior performance.

\noindent\textbf{ResNet-56 on CIFAR-10}.
As summarized in Table~\ref{tab:comparision_cifar10}, compared to LRPET~\cite{guo2022compact}, which led to a 0.23\% drop in terms of the Top-1 accuracy, our FSCL achieved a Top-1 increase by 0.39\% and a higher pruned rate in FLOPs. With more reduction in FLOPs (54.4\%), our FSCL could achieve a 0.26 Top-1 accuracy increase. But with lower FLOPs reduction (52.6\%), FPGM~\cite{he2019filter} and Pruning Criterion~\cite{9156434} result in Top-1 accuracy degradation. FSCL accomplished outstanding results.

\noindent\textbf{GoogLeNet on CIFAR-10}.
As shown in Table~\ref{tab:comparision_cifar10}, FSCL achieved the lowest accuracy drops compared to other baselines. GAL-ApoZ is the result of ApoZ obtained by the implementations of GAL~\cite{Lin2019TowardsOS}. Specifically,  FSCL pruned GoogLeNet with 75.3\% FLOPs reduction and 66.3\% parameters reduction only loses 0.02\% top-1 accuracy. These results showed  that our method could reduce the complexity of the model with inception modules.

\begin{table}[!t]
% 	\begin{minipage}[t]{0.48\textwidth}
% 	\renewcommand{\arraystretch}{1.0}
	\caption{Performance comparisons of ResNet-50 on ImageNet. "-" denotes that the results are not reported. $\downarrow$ is the accuracy gap between the baseline and the pruned model.}
	\label{tab:comparision_imagenet}
	\centering
	\scalebox{0.65}{
		\begin{tabular}{c|cc|cc}
			\hline
			Methods &\makecell[c]{P  Top-1(\%)/$\downarrow$(\%) }&\makecell[c]{P Top-5(\%)/$\downarrow$(\%)} &FLOPs/$\downarrow$(\%) &Params/$\downarrow$(\%) \\
			\hline
			CP~\cite{He2017ChannelPF} &72.30 / \,\,\,\,-\,\,\,\, &90.80 / 1.40  &\,\,\,\,\,\,-\,\,\,\,\,\, / 33.30 &\,\,\,\,\,\,-\,\,\,\,\,\,\, / \,\,\,\,\,\,-\,\, \\\
		    SSS-32~\cite{SSS2018} &74.18 / 1.94 &91.91 / 0.95  & 2.82B / 31.05 & 18.60M / 27.06 \\\
			NISP~\cite{Yu2018NISPPN} &\,\,\,\,\,\,-\,\,\,\,\,\, / 0.89 &\,\,\,\,\,\,-\,\,\,\,\, / 0.81  &\,\,\,\,\,\,-\,\,\,\,\,\, / 44.01 &\,\,\,\,\,\,\,\,\,-\,\,\,\,\,\,\,\, / 43.82 \\\
			GBN~\cite{zhonghui2019gate}  &75.18 / 0.67	&92.41 / 0.26	 &1.85B / 55.06 &\,\,\,\,\,\,\,\,\,-\,\,\,\,\,\,\,\, / 53.40   \\\
			
			FPGM~\cite{he2019filter}  &74.83 / 1.32	&92.32 / 0.55	 &\,\,\,\,\,\,-\,\,\,\,\,\, / 53.50 &\,\,\,\,\,\,-\,\,\,\,\,\,\, / \,\,\,\,\,\,-\,\,   \\\
			Hinge~\cite{Li2020GroupST}  &74.70 / 1.40	&\,\,\,\,\,\,-\,\,\,\,\, / \,\,\,\,-\,\,\,\,	 &\,\,\,\,\,\,-\,\,\,\,\,\, / 53.45 &\,\,\,\,\,\,-\,\,\,\,\,\,\, / \,\,\,\,\,\,-\,\,   \\\
			HRank~\cite{lin2020hrank}  &74.98 / 1.17	&92.33 / 0.54	 &2.30B / 43.77 &16.15M / 36.67   \\\
		
			LeGR~\cite{Chin2020TowardsEM} &75.70 / 0.40 &92.70 / 0.20  & 2.39B / 42.00 &\,\,\,\,\,\,-\,\,\,\,\,\,\, / \,\,\,\,\,\,-\,\, \\\
			AACP~\cite{lin2021aacp}  &75.46 / 0.48	&\,\,\,\,\,\,-\,\,\,\,\, / \,\,\,\,-\,\,\,\,	 &\,\,\,\,\,\,-\,\,\,\,\,\, / 51.70 &\,\,\,\,\,\,-\,\,\,\,\,\,\, / \,\,\,\,\,\,-\,\,   \\\
			DCFF~\cite{lin2021training}  &75.18 / 0.97	&92.56 / 0.50	 &2.25B / 45.30 &15.16M / 40.70   \\\
			LRPET~\cite{guo2022compact}  &74.38 / 1.77	&92.03 / 0.84	 &1.90B / 53.60 &12.89M / 49.50   \\\
			\textbf{FSCL(Ours)} &\textbf{75.84 / 0.31} &\textbf{92.79 / 0.08} &\textbf{1.80B / 55.99} &\textbf{11.78M / 53.80 }  \\
			\hline
	    \end{tabular}
	    
	} 
% \end{minipage}
\end{table}
% \vspace{-0.5cm}

% \subsection{ImageNet Results}
\noindent\textbf{ResNet50 on ImageNet}.
For ImageNet ILSVRC-12, we adopted ResNet50 for evaluation. We reported the results in Table~\ref{tab:comparision_imagenet}. FSCL consistently outperformed the counterparts on the FLOPs reduction and the parameter reduction. FSCL pruneed 53.80\% parameters to achieve 55.99\% FLOPs reduction with only 0.31\%/0.08\% Top-1/5 accuracy drop. Compared with those methods, FSCL achieves state-of-the-art performance, which shows that our FSCL can identify the redundant filters, and it is  effective on large-scale datasets.

\vspace{-0.2cm}

\subsection{Ablation Study}

\begin{figure}[!h]
	\centering
	\includegraphics[width=0.85\columnwidth]{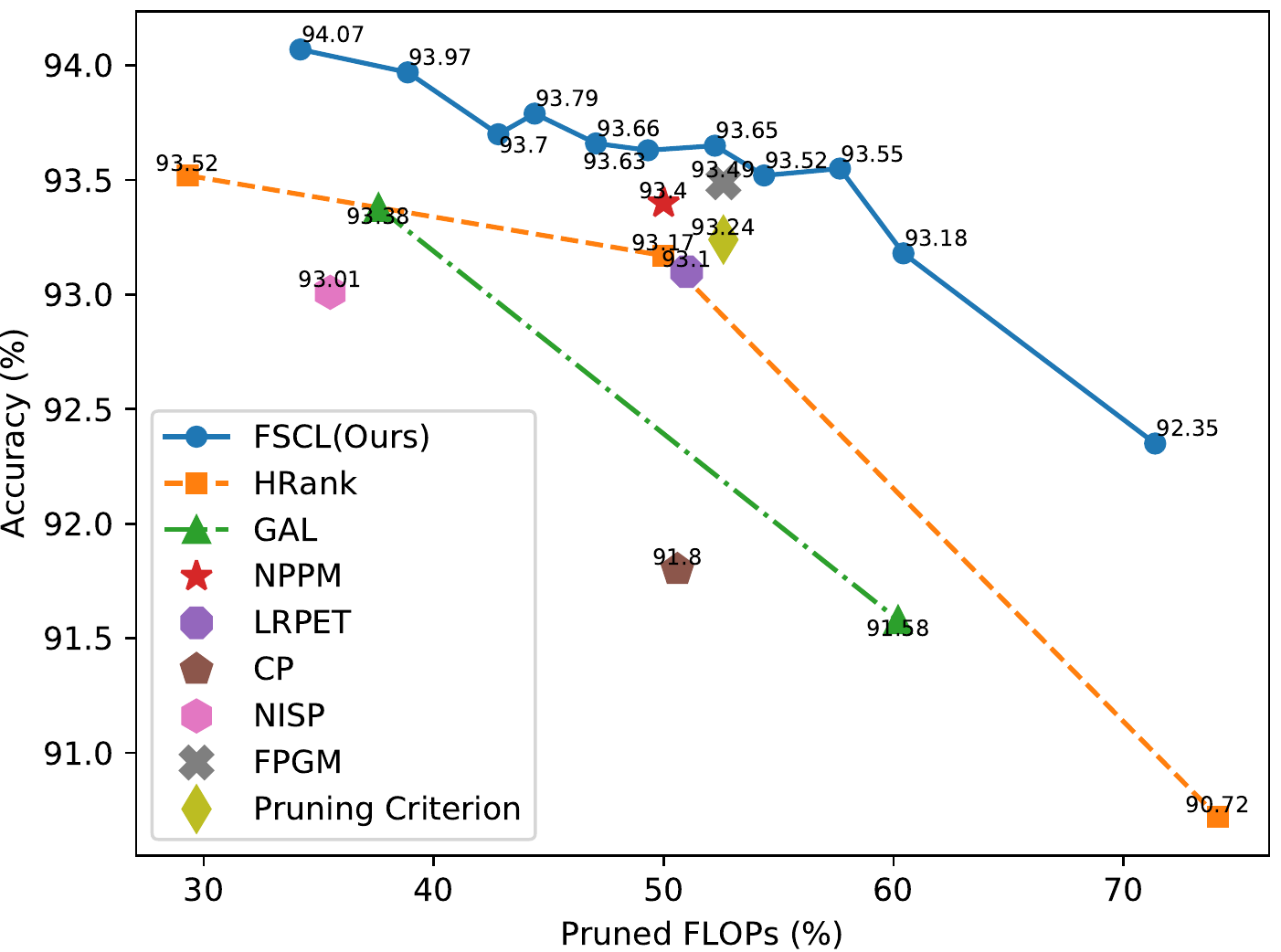}
	\caption{Accuracy of ResNet-56 using various pruned FLOPs. The dotted orange line indicates the accuracy of the baseline (unpruned ResNet-56).}
	\label{fig:flops}
\end{figure}

\textbf{Varying Pruned FLOPs}.
We performed experiments of FSCL to explore the relationship between pruned FLOPs and the accuracy of ResNet-56 ( Figure~\ref{fig:flops}). Compared to the original model, FSCL could reduce over 55\% of FLOPs without loss in accuracy. FSCL has a regularization effect on the neural network. Compared to others with the same baseline, we got higher accuracy on similar pruned FLOPs.

\vspace{-0.3cm}

\vspace{-0.5cm}
% conclusion
\section{Conclusion}

In this paper, we analyze the relationship of filters in consecutive convolutional layers and reveal that the filters' similarity can be utilized to slim the model. So we propose a novel filter pruning method called FSCL, which ranks the filter importance by their similarity in consecutive layers. FSCL can detect the filters whose corresponding features are more worthless for model compression. Extensive experiments on various modern CNNs show that our intuitive and effective FSCL achieves state-of-the-art performance.

\bibliographystyle{IEEEbib}
\bibliography{IEEEbib}
% \bibliography{strings,refs}

\end{document}